\documentclass[energies,article,accept,pdftex,moreauthors]{Definitions/mdpi} 

\firstpage{1} 
\pubvolume{16}
\issuenum{6}
\articlenumber{2837}
\pubyear{2023}
\copyrightyear{2023}
\externaleditor{Academic Editors:  Luis Roman and Mona Faraji Niri}
\datereceived{1 March 2023} 
\daterevised{15 March 2023} 
\dateaccepted{16 March 2023} 
\datepublished{18 March 2023} 
\hreflink{https://\linebreak doi.org/10.3390/en16062837} 

\Title{To Charge or to Sell? EV Pack Useful Life Estimation via LSTMs, CNNs, and Autoencoders}

\TitleCitation{To Charge or to Sell? EV Pack Useful Life Estimation via LSTMs, CNNs, and Autoencoders}

\Author{{Michael Bosello} 
 $^{1,2,}$*\orcidA{}, Carlo Falcomer $^{1}$\orcidB{}, Claudio Rossi $^{3}$\orcidC{} and Giovanni Pau $^{1,2,4}$\orcidD{}}

\AuthorNames{Michael Bosello, Carlo Falcomer, Claudio Rossi and Giovanni Pau}

\AuthorCitation{Bosello, M.; Falcomer, C.; Rossi, C.; Pau, G.}

\address{%
$^{1}$ \quad Department of Computer Science and Engineering, University of Bologna, 40126 Bologna, Italy\\
$^{2}$ \quad Autonomous Robotics Research Center, Technology Innovation Institute (TII), \linebreak 
Abu Dhabi P.O. Box 9639, United Arab Emirates\\
$^{3}$ \quad Department of Electrical, Electronic and Information Engineering, University of Bologna, \linebreak 40136 Bologna, Italy\\
$^{4}$ \quad Samueli Computer Science Department, University of California, Los Angeles, CA 90095, USA}

\corres{Correspondence: michael.bosello@unibo.it}

\abstract{
  Electric vehicles (EVs) are spreading fast as they promise to provide better performance and comfort, but above all, to help face climate change. Despite their success, their cost is still a challenge.
    Lithium-ion batteries are one of the most expensive EV components, and have become the standard for energy storage in various applications.
    Precisely estimating the remaining useful life (RUL) of battery packs can 
    encourage 
    their reuse and thus help to reduce the cost of EVs and improve sustainability. A correct RUL estimation can be used to quantify the residual market value of the battery pack. The customer can then decide to sell the battery when it still has a value, i.e., before it exceeds the end of life of the target application, so it can still be reused in a second domain without compromising safety and reliability.
    This paper proposes and compares two deep learning approaches to estimate the RUL of Li-ion batteries: LSTM and autoencoders vs. CNN and autoencoders. The autoencoders are used to extract useful features, while the subsequent network is then used to estimate the RUL. Compared to what has been proposed so far in the literature, we employ measures to ensure the method's applicability in the actual deployed application. Such measures include (1)~avoiding using non-measurable variables as input, (2) employing appropriate datasets with wide variability and different conditions, and (3) predicting the remaining 
    ampere-hours 
    instead of the number of cycles. The results show that the proposed methods can generalize on datasets consisting of numerous batteries with high variance.
}


\begin{document}

\section{Introduction and Background}

Electric vehicles (EVs) are becoming central to the automotive industry as they can address current automotive limits. Their constant growth is due to their improved performance and efficiency, but especially for their suitability in addressing environmental challenges, i.e., urban pollution and global warming \cite{HOFMANN2016995, ZOU201625}. Internal-combustion-based vehicles contribute to global carbon emissions by 14\% of the total \cite{epa2014}; thus, they are facing restrictions in leading markets that aim to  reduce their environmental footprint \cite{ZOU201625, eu2012}. Internal-combustion-based vehicles are also a prominent source of artificial fine particulate matter ($PM_{2.5}$) \cite{es981276y, kheirbek2016contribution}. Air pollution is one of our greatest social issues since it has a severe impact on health and society \cite{koolen2019air}, possibly causing different diseases and even premature death \cite{anderson2005ambient, brunekreef2005epidemiological}. EVs are a milestone in addressing such challenges to humanity as they can potentially remove personal transportation from the environmental impact equation.

A core component of EVs is the battery. Lithium-ion (Li-ion) batteries have become the standard for energy storage in EVs \cite{OPITZ2017685, MANZETTI20151004}. They have several advantages compared to traditional batteries such as lead-acid or nickel-metal hydride: high energy and power density, low self-discharge, environmental adaptability, long lifetimes, and \mbox{high reliability \cite{HANNAN2017834, ZHANG2019100951}.}
These advantages have led to the wide use of Li-ion batteries in EVs and in several safety-critical areas such as space applications \cite{liu2014data}, aircraft, and backup energy systems.
The safety and reliability of Li-ion batteries are critical concerns for such applications \cite{hu2017condition}. 
Li-ion batteries are employed in safety-critical areas, so their defects can cause fatal system failures. For example, various Boeing 787 
aircraft 
caught fire because of Li-ion battery malfunctions \mbox{in 2013 \cite{en6094682},} and NASA lost a spacecraft because of the lack of power supply due to a false battery over-charging indication in 2006 \cite{nasa2007}. Such high-impact failures have also recently appeared in the EV domain, with well-known manufacturers recalling hundreds of thousands of EVs due to fire risk \cite{ValdesDapena2021May, Hawkins2020Nov}. 
Another far more significant challenge for Li-ion batteries is their cost. While EVs are promising on various fronts, their cost is still a considerable drawback \cite{7112507}, and the battery is one of the most expensive components of EVs \cite{MAHMOUDZADEHANDWARI2017414}.

The design of an appropriate battery management system (BMS) is crucial to reducing costs and increasing vehicle efficiency and security \cite{8986572, JOHNSON2014582}. One of the major tasks of the BMS is to evaluate the current health conditions of the battery as it degrades over time. This degradation is an irreversible process related to the repetitive charging and discharging operations and electrochemical reactions inside the battery \cite{BARRE2013680}. Predominant indicators are battery capacity and internal resistance, which inform us about the battery residual energy and power capabilities, respectively \cite{XIONG201818}, indicated by the state of health (SOH). The SOH and the remaining useful life (RUL) are the most crucial parameters of battery health that must be estimated by the BMS \cite{HANNAN2017834}. 
The SOH quantifies the deterioration level compared to a brand new battery. While it has not been formally defined by industry \cite{LU2013272}, it is typically expressed through a percentage of capacity loss or power loss (increase in battery
resistance) \cite{FAN2020101741, 9036949}. We will consider the capacity loss (SOHc), which is defined by 
\begin{equation} \label{eqn:soh}
    SOH = \frac{C_t}{C_0}\cdot100(\%)
\end{equation}
where $C_t$ is the current capacity and $C_0$ is the nominal capacity.
\textls[-15]{The International Electrotechnical Commission (IEC) \cite{IEC}, International Organization for Standardization (ISO)~\cite{ISO},} and Institute of Electrical and Electronics Engineers Standards Association (IEEE-SA) \cite{IEEE-standard} have proposed standards to measure the battery capacity in a standard condition using direct methods that are taken as a reference to compute $C_t$ and $C_0$. 
The BMS adjusts its functioning according to the estimated SOH to ensure vehicle performance and safety until the health indicators reach the target limits, after which the battery should be replaced. Battery manufacturers usually set the capacity threshold under which the battery is no longer suitable for EV applications to 80\% of the nominal capacity \cite{DUONG2000244}, measured under a standardized test. Such threshold is called the end-of-life (EOL). Despite this, the battery might be replaced before the threshold if the internal resistance rises above a \mbox{normal level \cite{XIONG201818}.}
The threshold is also recommended to be 80\% by the Center for Advanced Life Cycle Engineering (CALCE) at the University of Maryland \cite{calce} and 70\% by NASA’s Prognostics Center of Excellence (PCoE) \cite{nasadata}.
In the context of replacement and secondary use planning, it is helpful to \emph{{predict} 
} how the SOH will evolve through time and when the battery will reach its EOL. This is defined by the RUL, which is typically described as the number of cycles remaining until EOL \cite{chen2020remaining, en10050691, NG2014114}.

A robust SOH estimation by the BMS is fundamental to ensure battery reliability as well as prevent failures and hazards \cite{8986572, HU2019109334}, but also to determine the acceleration performance and the driving range of the EV \cite{SONG2020101836, en12224338}{,} necessary for a pleasant driving experience{,} 
and finally to quantify the residual market value of the batteries \cite{AHMADI201464}.
However, the correct estimation of the RUL encourages the reuse of batteries, as removing the battery before it exceeds the end of life of the target application allows reuse in a second domain without compromising safety and reliability \cite{SHEN2019100817}.
Batteries can therefore be employed in secondary applications with lower power requirements.
This can have a significant impact in terms of both sustainability and market value \cite{harper2019recycling}.
With the growing number of new EVs, the waste produced by the spent battery packages is also increasing. Their recycling processes can have a considerable economic and environmental impact. The \linebreak interested reader is referred to \cite{prazanova} for lithium-ion battery recycling in the EV context.
To summarize, improving the estimation of SOH and RUL contributes to the spreading of EVs in two ways: (1) by ensuring security and reliability, (2) by reducing costs and waste through battery reuse.

\begin{figure}[H]
    \begin{adjustwidth}{-\extralength}{0cm}
    \centering
    \includegraphics[width=0.82\linewidth]{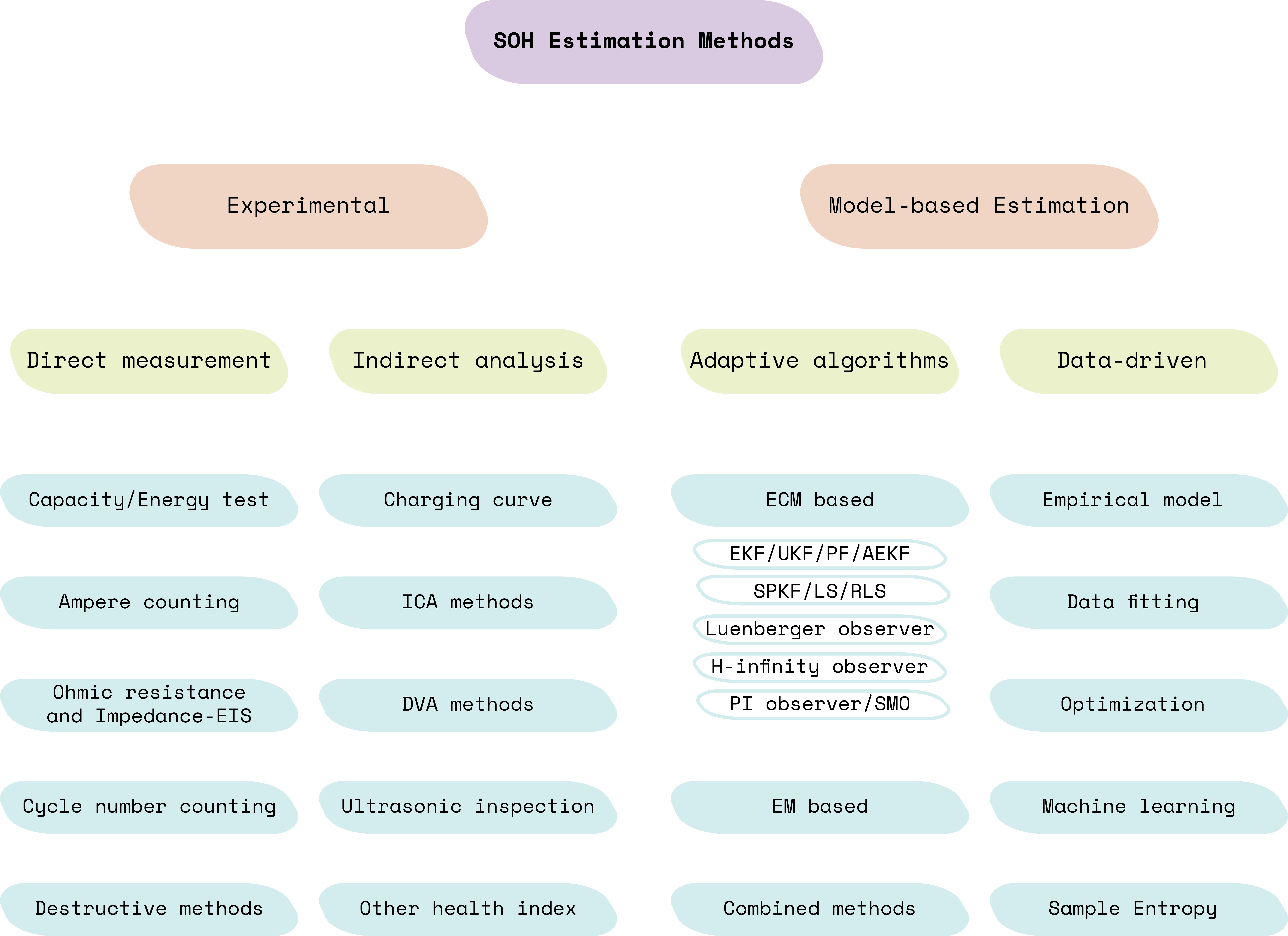}
    \end{adjustwidth}
    \caption{Classification of battery SOH estimation {methods.} 
    }
    \label{fig:sohmethods}
\end{figure}

The estimation of the SOH and its prediction (the RUL) is a challenging task. The capacity of a cell cannot be directly measured, so indirect measurements are used instead by using related variables.
The SOH 
can be precisely computed in laboratory conditions, but it significantly differs from the working conditions of real applications \cite{XIONG201818}. This, unfortunately, does not apply to the real-world EVs that have to employ online estimation \linebreak algorithms \cite{8986572}.
Battery aging involves many variables, such as charge/discharge current, voltage, and operating temperature.
EV batteries' working conditions are also highly dynamic as they change with the environment and the user's driving style \cite{SONG2020101836}.
As a result, it is challenging to design accurate physical models due to complex degradation mechanisms and operations. Furthermore, it requires much knowledge about the phenomena involved and experimental data acquired in controlled situations, which could be unavailable or quite expensive to collect \cite{SHEN2019100817}.
SOH estimation techniques can be classified into two macro-categories: experimental and model-based estimation methods \cite{8986572, XIONG201818}. Experimental methods analyze aging behavior through numerous laboratory tests. As mentioned above, this is typically not achievable on board due to the required equipment and the dynamic driving context.
Model-based methods can be further divided into adaptive algorithms and data-driven approaches, and the latter also includes RUL prediction. Adaptive algorithms use mathematical models and numerical filters (e.g., equivalent circuit model and Kalman filters). In contrast, data-driven methods use black box models, which find the mapping between the input and the target. Figure \ref{fig:sohmethods} summarizes the main categories of SOH estimation methods.
The following section will focus on machine-learning-based SOH estimation and RUL prediction techniques and their advantages. For a detailed review of the other methods for SOH estimation, please refer to \cite{XIONG201818}, and for RUL prediction, please refer to \cite{LIPU2018115}.

This paper contributes in several directions to the applicability of deep-learning-based RUL estimation in practical applications.
\begin{itemize}
\item We propose a novel RUL definition in the machine learning 
context, 
based on ampere-hours, to push forward the applicability to real cases. The first application of deep learning techniques on an RUL that is not based on the simplified concept of cycles is also provided.
\item Two models for RUL estimation are presented and compared on the NASA Randomized dataset: autoencoder plus CNN and autoencoder plus LSTM. In addition, an LSTM is proposed to predict the RUL in the UNIBO Powertools dataset. The results show that the 
particular 
autoencoder can effectively extract the relevant features of the cycle curves, while the CNNs and the LSTMs can be used to estimate the RUL.
\item Two vast datasets containing batteries cycled with an extensive set of different conditions and variables are used in the experiments to ensure generalization. Compared to the data used in the literature so far (with a limited amount of batteries typically discharged under constant current), the examples used in this paper present many more batteries and conditions that are more challenging to predict. All the relevant details about the data selection and splitting are detailed, ensuring transparency in the results.
\end{itemize}

The paper is structured as follows. The next section provides an overview of the deep-learning-based techniques used for RUL estimation and the problems that affect their deployment. The relevant works are also summarized. In Section \ref{sec:experiments}, the datasets and the models used are described. Section \ref{sec:results} shows and compares the results obtained using the models. Finally, Section \ref{sec:conclusions} concludes the paper with a summary of the results and future research directions.

\section{Related Works} \label{related}

The recent success of machine learning in several domains and the availability of data and computing power have motivated the development of novel methods for battery state estimation. Data-driven battery state estimation methods are becoming increasingly popular \cite{9036949}.
In particular, the attention to using deep learning (DL) for battery status estimation has increased over time.
Data-driven methods provide several advantages~\cite{8986572}. They allow us to obtain better results in real complex applications, as complete knowledge about degradation mechanisms is still lacking. They do not require expert knowledge about the degradation phenomena as they only rely on enough operational data from which key features are extracted. They are also suitable for execution on hardware with limited capabilities compared to adaptive algorithms that are more computationally demandin~\mbox{\cite{en12224338, 8536873}}. While the training phase is demanding as well, the execution is efficient and can run on BMS hardware, with inference models in the order of a few hundred megabytes \cite{NIPS2012_4824}.
Last but not least, data-driven methods enable the prediction of the SOH (i.e., the RUL), while other techniques are typically limited to estimating the current SOH.

Drawbacks are present, but the benefits compensate for them. The main 
drawbacks 
 are limited interpretability and inaccessibility to physical parameters (e.g., internal \linebreak resistance) \cite{7811778}.
Before proceeding, it is worth noting that we may have a conflict of terms. While, in the context of deep learning, ``prediction'' typically indicates the result of a neural network, in the context of signal processing, it means predicting the future value of a time series. In this paper, we will use the term \emph{estimation} to indicate the estimation of the current SOH. In contrast, we will use the word \emph{prediction} to indicate the expected RUL, as it can be conceived as the prediction of the future SOH.


In the last two years, numerous works using DL for SOH and RUL estimation have been proposed in the literature.
In the following, the common approaches (and issues) in the various papers are first presented to avoid repetition, and then the single articles \mbox{are analyzed.}

The variables measured by the BMS are usually voltage (V), current (I), and temperature (T). Such variables are sampled at high rates during subsequent charge and discharge cycles, resulting in very long time series.
The variables most used as input are voltage, current, and temperature, as they can capture the battery aging factors \cite{en12224338}, but sometimes the sampling time and the state of charge (SOC; i.e., the charge level) are also used. In the case of the RUL, the SOH itself has also been used as input.
It is recalled, however, that care must be taken when non-measured variables (i.e., SOC, SOH) are employed, as errors could accumulate and robust estimation of the input variable might not always be available---
this is the 
first issue affecting applicability.
The variables can originate from the charging cycles, the discharging cycles, or both.
The resulting time series are often presented to the network through a moving window, i.e., the NN makes the estimation based on the set of features at time $t$ plus their last $N$ values.
A popular approach is to use a recurrent neural network (RNN)---in one of its various forms---to find the relation between the SOH or RUL and the time series.
Long short-term memory (LSTM) \cite{hochreiter1997long} networks are 
particular 
RNNs that are able to handle long-term sequences and have become the baseline of recurrent networks. LSTMs and their variants are thus also widely used in the battery context.
Some experiments try instead to use convolutional neural networks (CNNs) to process time series, or to use a simple feed-forward NN (FFNN) preceded by specific pre-processing.
The literature offers several battery datasets to conduct such experiments.
One of the most used datasets is the NASA ``Battery Data Set'' \cite{nasadata}. It was the first battery dataset that was publicly available, and thus it has had a significant impact in the \mbox{field \cite{DOSREIS2021100081}}. The dataset consists of 34 Li-ion 18,650 cells cycled at various ambient temperatures; however, it includes only constant current (CC)-cycled batteries. Even though the dataset contains 34 batteries, the most common approach is to use a specific subset of three or four batteries. In 2014, NASA released another dataset (``Randomized Battery Usage Data Set'') \cite{bole2014randomized} also containing batteries cycled with a random current. A review of battery datasets is available in \cite{DOSREIS2021100081}.
The second issue to be addressed to ensure the suitability of the SOH and RUL methods to \emph{real scenarios} is the quality of the dataset used during testing. Most works use \emph{simplified databases} with batteries cycled under CC discharge, a condition not applicable to EV operation.
Another applicability obstacle in the case of RUL is in the \emph{definition of RUL} itself. As already mentioned, RUL is defined as the remaining cycles before EOL. In the EV context, we have partial charge and discharge cycles (e.g., discharge to 40\%, charge to 80\%, discharge to 30\%, and so on) as the vehicle can be recharged starting from different SOCs and can be unplugged before the full charge. An equivalent full cycle (0\% to 100\%) has little practical meaning, therefore the definition of RUL has to be rethought. A valid candidate to represent the RUL in the EV setting is the remaining ampere-hour (Ah) value that the battery can deliver before reaching EOL.
The measures to prevent applicability pitfalls are then (1) avoiding using non-measurable variables as input, (2) employing appropriate datasets with wide variability and different conditions, and (3) not using cycles to define the RUL.

The estimation of the SOH has recently become quite robust as it has been applied to realistic datasets. SOH estimation is well established on simplified datasets \cite{8986572, 9036949, onlineestimationrnn2020}; here, we report the recent advances on complex datasets.
In \cite{FAN2020101741}, a hybrid network comprising a gated recurrent unit (GRU; a well-known variation of LSTM) and a CNN is used for SOH estimation.
The inputs are the raw V, I, and T data of the charging curve, converted to a fixed size history of 256. The input converges 
 toward the two parallel streams (GRU and CNN) concatenated in the last layer.
A maximum estimation error of 4.3\% on the Randomized NASA dataset is reported.
The authors of \cite{en12224338} proposed an SOH estimation method based on the Independently RNN (IndRNN) and tested it on the Randomized NASA dataset. Here, a discharge cycle is represented by 18 features, including average V, I, and T, as well as the capacity, the time elapsed, and the time periods of each current load. While it achieves superior results, whether it will work in real applications needs to be clarified as it also takes capacity as input. In the experiments, the capacity in input is calculated. In contrast, the proper way to conduct the investigation should have used the capacity estimated by the network itself in the previous time step.
In \cite{SHEN2019100817}, a CNN takes as input the V, I, and capacity of the charging cycle discretized in 25 segments. The output is the capacity computed on the corresponding discharge cycle. Both capacities are computed 
by coulomb counting. 
Applicability is at least doubtful in this case too.
In \cite{SONG2020101836}, a private database of real-world data collected from 700 vehicles (full-electric or hybrid) is used to train an FFNN for SOH estimation. The parameters employed are the accumulated mileage of cars, the C-rate distribution, the SOC range (the SOC is divided into five ranges, and the SOC range indicates such range), and cell temperatures. The number of variables is reduced to a lower dimensionality using principal component analysis (PCA). 
The results are impressive, with a maximum error of 4.5\% and RMSE of 1.1\%, which become 2.2\% and 0.45\% if considering only fully electric vehicles.

Moving to RUL estimation, it has not yet reached the robustness and applicability of SOH estimation. The experiments described in the literature are limited to oversimplified datasets that present only CC-cycled batteries. While adequate performances are typically achieved, there are still some critical issues regarding data quality and applicability. 
In \cite{9037249}, a temporal convolutional network (TCN) produces RUL estimations. The input is the history of the SOH, processed through a moving window. Tests are performed on three CC batteries from the NASA dataset and two CC batteries from the CALCE dataset. As it uses the history of SOH, a robust SOH estimation is necessary to ensure its applicability. As the experiments rely on ground truth SOH, it needs to be clarified if the proposed method will have the same performance using estimated SOH levels that are thus affected by some error. In addition, a long warm-up is required, as the first output is produced after a minimum $t$ of 30 cycles.
Both \cite{8079316} and \cite{ZHANG2020108052} propose using an LSTM network that takes the SOH's history as input. In the first case, the output is the RUL; in the second case, it is the k-step ahead SOH, 
which can be reduced to the RUL. 
The datasets used are one Panasonic CC battery and four CC batteries from the NASA set, respectively. Both works present the same issues as the TCN-based one (here, the warm-up is even longer).
The most promising article is \cite{LI2020228069}, which presents a variant of LSTM, namely AST-LSTM.
Two AST-LSTMs are trained, one for estimating the SOH and one for the RUL. The input of the first model includes V, I, T, and the sampling time of the discharge cycles. The second model uses instead the history of SOH estimated from the previous one. As the 
SOH input 
 is estimated, the RUL approach is also suitable for real scenarios. Experiments are conducted on 12 batteries from the NASA dataset. The approach still needs to be tested on better datasets, and the warm-up is too long.
\mbox{In \cite{ZHANG2019100951}, }the IC discharge curve is computed from the V, I, and sampling time. The features extracted from the curve are inputted to a small NN that estimates the SOH and RUL. This method is, however, applicable only to CC discharging. In addition, it has a high computation complexity and low performance.
\mbox{In \cite{8418374},} an autoencoder is used to perform dimensionality reduction starting from the V, I, T, and sampling time of both charge and discharge cycles, plus the capacity estimated during discharge. In addition to the applicability doubts, the accuracy is so low that the approach is substantially inapplicable. Another autoencoder approach from the same main author was proposed in \cite{9137406}. Here, the autoencoder is used instead to augment the data dimension, and then the result is processed by two branches: an LSTM and a CNN. The features extracted by them are concatenated and fed to the final NN that returns the RUL prediction. While the performance has improved, the dataset used is still insufficient, as only CC-discharged batteries are considered.
The properties of the above-mentioned works are summarized in Table \ref{table:rul}.

\startlandscape
\begin{table}[H]
\caption{Summary of current DL-based RUL estimation approaches.}
\begin{tabularx}{\textwidth}{p{4.0cm}p{3.2cm}p{1.7cm}p{3.5cm}p{3.6cm}p{8.55cm}} 
     \toprule
     \textbf{Method} & \textbf{Input} & \textbf{Output} & \textbf{Dataset} & \textbf{Performances} & \textbf{Issues} \\ 
     \midrule

     TCN \cite{9037249} & History of SOH & 
     RUL &
     NASA, 3 CC batteries
    
    (\#5, \#6, \#18)

     CALCE, 2 CC batteries 
     
     (\#CS\_34, \#CS\_35) & RMSE up to 0.048 &
     \vspace{-0.43cm}
     \begin{itemize}
        \item It requires a robust SOH estimation
        \item It is not clear if it will work if SOH is affected by errors
        \item Insufficient dataset variability
        \item Long warm-up (minimum starting cycle for NASA is 30 and the starting cycle for CALCE is 360)
     \end{itemize}
      \\

     LSTM network \cite{8079316} & History of SOH & RUL & Panasonic 18,650 
     
     (1 CC battery) & - & 
     \vspace{-0.43cm}
     \begin{itemize}
        \item It requires a robust SOH estimation
        \item It is not clear if it will work if SOH is affected by errors
        \item Insufficient dataset variability
        \item Long warm-up (start after 50\% of battery life)
     \end{itemize}
      \\

     LSTM network \cite{ZHANG2020108052} & History of SOH & K-step ahead SOH & 
     NASA, 4 CC batteries 
     
     (\#5, \#6, \#7, \#18) & MAE 1.92 
     
     (on battery \#5) &   
     \vspace{-0.43cm}   
     \begin{itemize}
       \item It requires a robust SOH estimation
       \item It is not clear if it will work if SOH is affected by errors
       \item Insufficient dataset variability
       \item Long warm-up (start at cycle 60)
    \end{itemize} 
    \\

    Variant of LSTM.
    
    SOH is estimated and then used to predict RUL \cite{LI2020228069} & V, I, T, sampling time (discharge) & SOH+RUL & NASA (12 batteries) &
   SOH: RMSE up to 0.059

   RUL: RMSE up to 0.026 
   
   (on battery \#5)
    
    &
    \vspace{-0.43cm}       
    \begin{itemize}
       \item Insufficient dataset variability
       \item Long warm-up (start at cycle 50)
    \end{itemize} \\

     Features are extracted from the IC discharge curve and given as input to two NNs~\cite{ZHANG2019100951} & V, I, sampling time 
     
     (discharge) & SOH+RUL & 
     
     NASA, 4 CC batteries
     
     (\#5, \#6, \#7, \#18) & SOH: MRER up to 1.25\%

     RUL: RMSE up to 5.41 & 
     \vspace{-0.43cm}
     \begin{itemize}
        \item Applicable only to CC discharging
        \item Computational complexity
        \item Insufficient dataset variability
        \item Low performances
     \end{itemize}
      \\

     Feature extraction with 
     
     autoencoder followed by a DNN \cite{8418374} & V, I, T, sampling time (charge + discharge), capacity, i.e., SOH 
     
     (discharge) & RUL & 
     NASA, 3 CC batteries
     
     (\#5, \#6, \#7) & RMSE up to 13.2\% 
     
     (on battery \#7) &
     \vspace{-0.43cm}
     \begin{itemize}
        \item It requires a robust SOH estimation
        \item It is not clear if it will work if SOH is affected by errors
        \item Insufficient dataset variability
        \item Insufficient performances
     \end{itemize} 
     \\

     Auto-CNN-LSTM: data augmentation with
     autoencoder followed by CNN+LSTM extractor \cite{9137406}
     & V, I, T, sampling time (charge + discharge)
     & RUL 
     & NASA, 6 CC batteries
     
     (\#5, \#6, \#7 \#25, \#27, \#28) & RMSE up to 5.03\% 
     
     (on battery \#7 and \#28) &
     \vspace{-0.43cm}
     \begin{itemize}
        \item Insufficient dataset variability
     \end{itemize} 
     \\

     \bottomrule
     
\end{tabularx}

\label{table:rul}
\end{table}
\finishlandscape

\section{Experiments}
\label{sec:experiments}
In this work, we propose and compare three models to predict the remaining useful life of Li-ion batteries. The contributions and improvements focus on the model's applicability to real-world scenarios and the transparency in defining batteries and methodologies used. Existing works have used limited datasets without specifying why only some batteries from the selected dataset have been used. They also conform to the definition of RUL based on cycles, which has little meaning in actual applications. Finally, most of them do not provide enough information about the data structure and how data are employed.
To achieve the desired aims, three aspects have been covered: input definition, output (RUL) definition, and the use of datasets with wide variability.

\begin{itemize}
  \item {\textbf{Input:}} 
 The only information used is voltage (V), current (I), and temperature (T), as using only measurable variables boosts applicability. The data are taken from the \emph{discharge} cycles and are organized in time series, in the format $[cycle, timestep, variable]$. As explained in Section \ref{models}, at each cycle $n$, the input given to the network is based solely on the cycle $n$ and the previous ones ($n-1, \ldots, n-N$; where $N$ is the history length), i.e., no information from the future is used.
  \item \textbf{Output---RUL definition:} As detailed in Section \ref{related}, defining the RUL as the number of remaining cycles has no practical meaning. Here, instead, the RUL is defined as the \emph{normalized remaining ampere-hour (Ah)} that the battery can deliver before reaching EOL. 
\end{itemize}

This can be named 
ah-RUL 
and is computed as:
\begin{adjustwidth}{-\extralength}{0cm}
  \begin{equation}
      ahRUL(n) = [trapz(current_{[:, :]}, time_{[:, :]}) - trapz(current_{[n:EOL, :]}, time_{[n:EOL, :]}) ] /nomcap
  \end{equation}
  \end{adjustwidth}
\noindent {where} 
 {$n$} 
 is the current cycle number, {$trapz$} is the approximate integral via the trapezoidal method, {$current$} is the matrix of currents where the x is the cycle and y is the timestep, {$time$} is the matrix of the elapsed time corresponding to the current measurement, and {$nomcap$} is the nominal capacity of the battery.

  As the RUL is a slowly changing value, predicting at every cycle (rather than at each timestep) is more than sufficient. Thus, for each cycle $n$ (and history $N$), the model predicts the ah-RUL at the current cycle. 

The code for the pre-processing, as well as the networks, the trained models, the results, and the plots, are available in the repository of the project, which can be accessed at \href{https://github.com/MichaelBosello/battery-rul-estimation}{https://github.com/MichaelBosello/battery-rul-estimation} {(accessed on 15 March 2023).} 

\subsection{Datasets}

Two datasets have been used. The NASA Randomized Battery Usage dataset \cite{bole2014randomized}, which is used to investigate the performances of batteries cycled with a random current, and the UNIBO Powertools Dataset, which was released by our team in \cite{10.1145/3462203.3475878}, and contains batteries with different specifications that were cycled under different conditions.

\subsubsection{NASA Randomized Battery Usage Dataset}

The NASA Randomized dataset consists of data from 28 batteries. The batteries are lithium cobalt oxide 18,650 cells
with a nominal capacity of 2.2 Ah. The cells were continuously operated by repeatedly charging and discharging them according to the corresponding profile. At every 50 cycles, a series of reference charging and discharging cycles were performed to provide battery health status. The batteries are split into \mbox{7 groups} containing 4 cells each according to the charge/discharge profile and temperature used. The randomized charge/discharge profiles can be as follows: random walk (RW), i.e., the selection of the current is random with a uniform distribution between the two current ranges; skewed RW, i.e., the current selection is random with a custom probability distribution skewed towards lower or higher currents. 

\begin{itemize}
    \item \textbf{RW1, RW2, RW7, RW8} batteries are repeatedly charged for a random duration between 0.5 and 3 h, then discharged to 3.2 V using a randomized sequence of currents between 0.5 A and 4 A. The discharge random profile is the RW. The setpoint is loaded every 5 min. Operated at room temperature.
    \item \textbf{RW3, RW4, RW5, RW6} batteries are repeatedly charged to 4.2 V and then discharged to 3.2 V using a randomized sequence of currents between 0.5 A and 4 A. The discharge random profile is the RW. The setpoint is loaded every 5 min. Operated at room temperature.
    \item \textbf{RW9, RW10, RW11, RW12} batteries are repeatedly charged and then discharged using a randomized sequence of currents between $-$4.5 A and 4.5 A. The charge and discharge random profile is the RW. The setpoint is loaded every 5 min. Operated at room temperature.
    \item \textbf{RW13, RW14, RW15, RW16} batteries are repeatedly charged to 4.2 V and then discharged to 3.2 V using a randomized sequence of currents between 0.5 A and 5 A. The random profile is the skewed high RW. The setpoint is loaded every 1 min. Operated at room temperature.
    \item \textbf{RW17, RW18, RW19, RW20} batteries are repeatedly charged to 4.2 V and then discharged to 3.2 V using a randomized sequence of currents between 0.5 A and 5 A. The random profile is the skewed low RW. The setpoint is loaded every 1 min. Operated at room temperature.
    \item \textbf{RW21, RW22, RW23, RW24} batteries are repeatedly charged to 4.2 V and then discharged to 3.2 V using a randomized sequence of currents between 0.5 A and 5 A. The random profile is the skewed low RW. The setpoint is loaded every 1 min. Operated at 40 {C} 
 temperature.
    \item \textbf{RW25, RW26, RW27, RW28} batteries are repeatedly charged to 4.2 V and then discharged to 3.2 V using a randomized sequence of currents between 0.5 A and 5 A. The random profile is the skewed high RW. The setpoint is loaded every 1 min. Operated at 40 C temperature.
\end{itemize}

\subsubsection{UNIBO Powertools Dataset}

The UNIBO Powertools dataset contains data from 27 batteries featuring various types of cells and experimental conditions collected in a laboratory test by an Italian equipment producer. Cells were charged at 1.8 A until 4.2 V and discharged at 5 A until $V_{eod}$. Capacity and resistance reference cycles were performed every 100 cycles. The batteries are split into 7 groups, according to the battery manufacturer, the cell type, the cell capacity, and the type of test.
The battery manufacturer is defined by a label, either D or E, to keep the brand name private. The cell type defines its intended use, and it can be M (mid-power), E (e-bike), or P (power tool). The tests performed on the cells are Standard (5 A discharge), High Current (8 A discharge), and Pre-conditioned (90 days' storage at 45 degrees C before testing). It follows the list of the groups, with the convention name \texttt{XW-Y.Y-AABB-T}, where X is the manufacturer, W is the cell type, Y.Y is the capacity, AABB is the delivery date (AA: week, BB: year), and T is the test type, followed by the list of cell numbers included in the group.

\begin{itemize}
    \item \textbf{DM-3.0-4019-S} 4 cells: 000, 001, 002, 003.
    \item \textbf{DM-3.0-4019-H} 3 cells: 009, 010, 011.
    \item \textbf{DM-3.0-4019-P} 7 cells: 013, 014, 015, 016, 017, 047, 049.
    \item \textbf{EE-2.85-0820-S} 4 cells: 006, 007, 008, 042.
    \item \textbf{EE-2.85-0820-H} 2 cells: 043, 044.
    \item \textbf{DP-2.00-1320-S} 8 cells: 018, 019, 036, 037, 038, 039, 050, 051 (039 has date 2420).
    \item \textbf{DM-4.00-2320-S} 2 cells: 040, 041.
\end{itemize}

\subsection{Models}
\label{models}

\subsubsection{NASA Randomized: AE-LSTM vs. AE-CNN}

For the NASA Randomized dataset, we propose and compare two networks. Both models use an autoencoder to compress the long time series in the set, in which timesteps range from 10k to 100k measurements per cycle. Using an autoencoder to perform the reduction allows us to retain most of the information, avoiding the loss of the fundamental information. 
The prediction of the ah-RUL is then made by passing the sequence of reduced cycles to a subsequent network, a CNN in one case and an LSTM in the other. 
CNNs are known to perform better on structured data with local information, which motivates their use in this use case.
We also considered LSTMs as they are designed to handle long-term sequences such as the battery aging process. Thus, they can learn the long-term degradation trend of batteries.

\begin{figure}[H]
    \begin{adjustwidth}{-\extralength}{0cm}
        \centering
        \includegraphics[width=\linewidth]{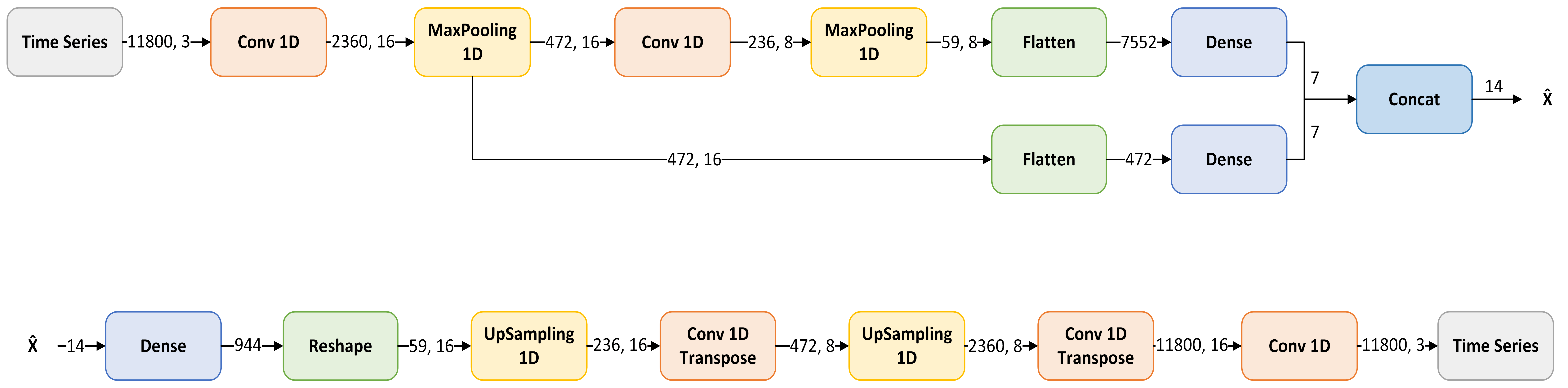}
    \end{adjustwidth}
    \caption{{The structure} 
 of the autoencoder used to compress the cycles. The skip link allows us to retain both local and global information}
    \label{fig:Autoencoder}
\end{figure}

As mentioned above, the autoencoder has to reduce the thousands of values per cycle to a small number of features (in the order of dozens) that are representative of the cycle.
This is done by an NN with an hourglass shape trained to copy its input to its output (in this case, the cycle measurements). As there is a bottleneck in the middle layer, it will learn a compact representation of the input that retains most information. To obtain the best reconstruction results, we employed a specific autoencoder structure \mbox{presented in \cite{MAGGIPINTO2018126}} that retains both \emph{local and global information}. The autoencoder structure is depicted in \mbox{Figure \ref{fig:Autoencoder}.} This network uses skip connections, i.e., jumps, to keep the features extracted not only from the last layer but also from the middle one. 

The encoder takes the time series of one cycle as input, composed of 11,800 measurements with 3 values each: voltage, current, and temperature. Voltage, current, and temperature are normalized between 0 and 1 using min-max normalization to prevent the magnitude of the value from affecting its importance. The min and max values are computed from the training set only. Depending on the experiment, the length of the time series in the group could be in the order of 10 k, 20 k, or 100 k. To reduce all the series to the same size, time series with 20 k measurements were sampled, keeping 1 value for each 2, 
and similarly, 
1 value for each 10 was kept for the 100 k time series. After that, the exceeding length was cut to 11,800.
The series goes into a 1D CNN with 16 filters, a kernel size of 10, and 5 strides. It follows max pooling with size 5. Here, the skip connection opens a fork: the features extracted so far are given to both a CNN layer and a flattening 
 layer followed by a dense layer. This dense layer with dimension 7 gives the first part of the encoded vector containing the local information. Coming back to the other path, the CNN has 8 filters with kernel size 4 and 2 strides, and is followed by max pooling with dimension 4, a flattening, and a dense layer with size 7 that produces the last part of the feature vector. The decoder performs the inverse operations mentioned above without the skip link. All the layers use the ReLu activation function. Adam is used to train the network with a learning rate of 0.0002, MSE loss, 500 epochs, and batch size 32.

Once the autoencoder has been trained, the encoder is used to reduce the cycles to a feature vector of size 14.
The feature vector is again normalized between 0 and 1, and then the time series for the RUL prediction is formed. The subsequent networks take a time series of $N=1000$ in the case of the CNN and $N=500$ in the case of the LSTM. The time series is formed by concatenating the vectors of the current cycle $n$ plus the previous $N$ cycles. If the current cycle number $n$ is below $N$, the time series is padded with zeros. A warmup of 15 cycles in training, and 30 in testing, is given to the network, i.e., the predictions start at cycles 15 and 30, respectively. The output of the nets, detailed at the beginning of the section, is also normalized between 0 and 1.

The CNN used has two 1D convolutional layers, with 64 and 32 filters, respectively, kernel sizes of 8 and 4, and  4 and 2 strides. It follows the flattening and three dense layers, with dimensions of 32, 16, and 1. All the layers use L2 kernel regularizers and ReLu activation, except for the last one, which uses 
linear activation.  
The Adam optimizer is used with a learning rate of 0.000003, Huber loss, 3000 epochs, and a batch size of 32.

The LSTM network has one masking layer to ignore padding zeros, two LSTM layers with 128 and 64 neurons each, and three dense layers with 64, 32, and 1 neuron each. All the layers again use L2 kernel regularizers but SELU activation, and linear activation in the last one. The same Adam optimizer is used with a learning rate of 0.000003 and Huber loss, but 500 epochs are performed.

\subsubsection{UNIBO Powertools: LSTM}

In the case of the UNIBO dataset, the autoencoder is not used since the cycle already has a low dimensionality. The length of the cycles of this dataset is reasonably short, in the order of 300 measures per cycle. In this case, using an autoencoder does not provide significant benefits in terms of dimensionality reduction, while it would add unnecessary complexity to the data processing pipeline.
Simpler pre-processing has been employed instead. In particular, the discharge cycle is reduced to six features: the average V, I, T, and their standard deviations.
The LSTM network used is the same as presented for the NASA Randomized dataset, considering both structure and hyperparameters. Likewise, the same history length, warmups, and normalization are employed.

\section{Results}
\label{sec:results}

\subsection{NASA Randomized}

Batteries from the different groups (as detailed in the previous section) have been used in both training and testing, to ensure reliable results. Six out of seven groups have been used. The group of batteries that have been cycled using the RW on both charge and discharge has not been used because it produced too much data to be handled (having a lot of micro-cycles). Plus, it can be considered an unrealistic use. Therefore, batteries RW9, RW10, RW11, and RW12 have not been used. Of the remaining 6 groups, each having 4 batteries, 3 batteries have been used for training and 1 battery for testing, with the following exceptions. The battery RW3 has been excluded because the temperature measurement is corrupted, and the battery RW20 has been excluded because all the measurements are 0 for almost all of the battery life. As a result, the groups of the batteries RW3 and RW20 had only 2 batteries used in training, making the learning even harder for such groups. The total number of batteries was 16 in training and 6 in testing. The complete list of the batteries used in training and testing is available in the project repository. The exact same splitting of batteries is used in the training and testing of both the autoencoder and the ah-RUL predictor, as it is assumed that at application time, neither of the networks will know the new data. 

To evaluate the performance of the autoencoder and the ah-RUL predictor, the average root-mean-squared error {(RMSE)} 
 is employed, as defined by the following formula:

\begin{equation}
  RMSE = \sqrt{\frac{\sum\limits_{i=1}^n (y_{i}-\hat{y}_{i})^2}{n}}
\end{equation}

\begin{figure}[H]
    \begin{adjustwidth}{-\extralength}{0cm}
        \centering
    \includegraphics[width=0.80\linewidth]{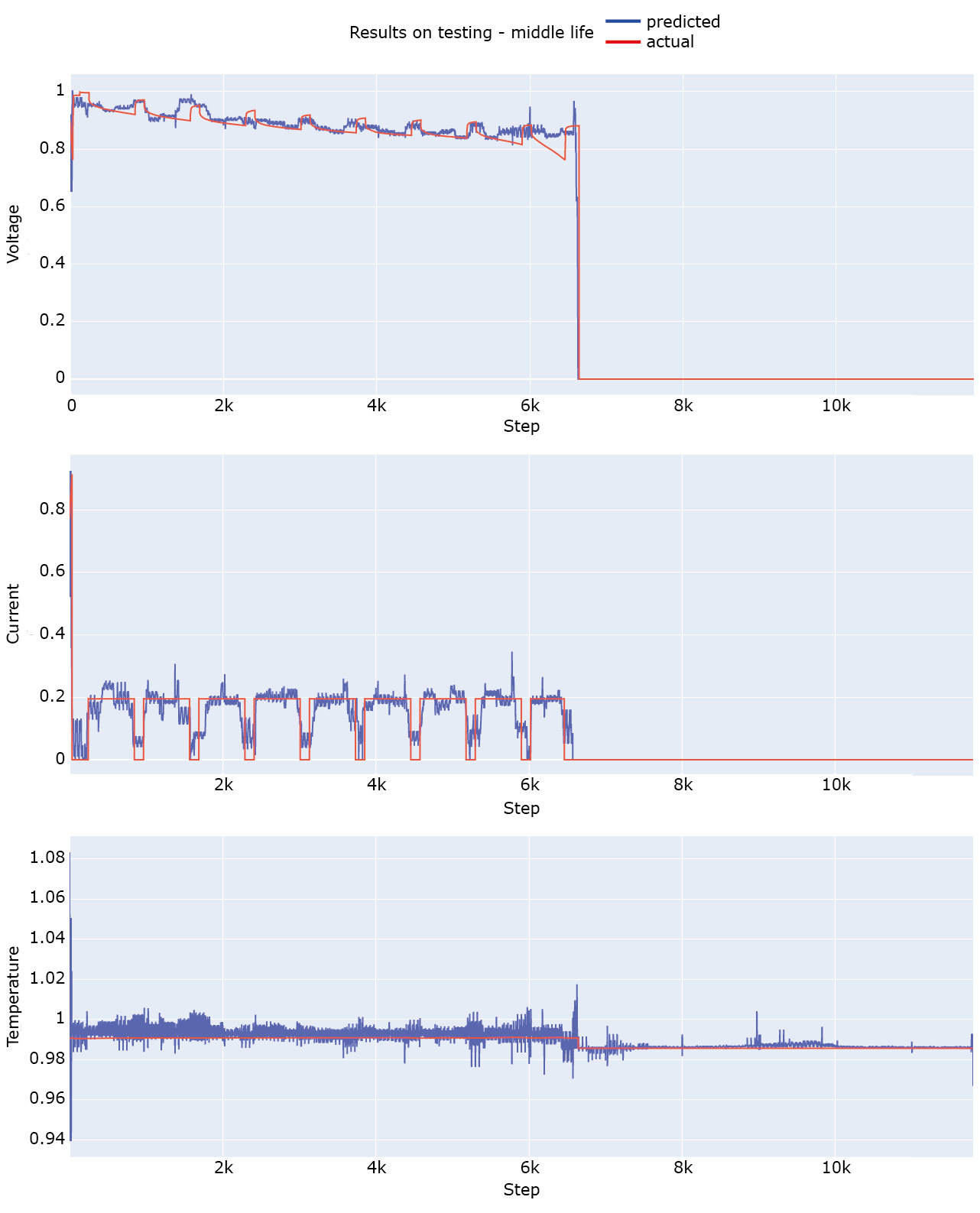}
    \end{adjustwidth}
    \caption{Example of the results of the autoencoder reconstruction capabilities on the testing set. The voltage, current, and temperature curves of a cycle in the battery's middle life were reconstructed from the extracted features.}
    \label{fig:Autoencoder-results}
\end{figure}

The average is computed on the RMSE obtained on each cycle of each battery. The reconstruction error of the autoencoder achieved a notable 0.0356 average RMSE on the testing set. The Autoencoder is thus able to retain the most important information of the curves at every phase of the battery life and cycled under different conditions, and then reconstruct them with high accuracy. An example of the original and reconstructed voltage, current, and temperature curves of a battery in its middle life is shown in Figure \ref{fig:Autoencoder-results}. 

\vspace{-8pt}

\begin{figure}[H]
 
    \includegraphics[width=0.96\linewidth]{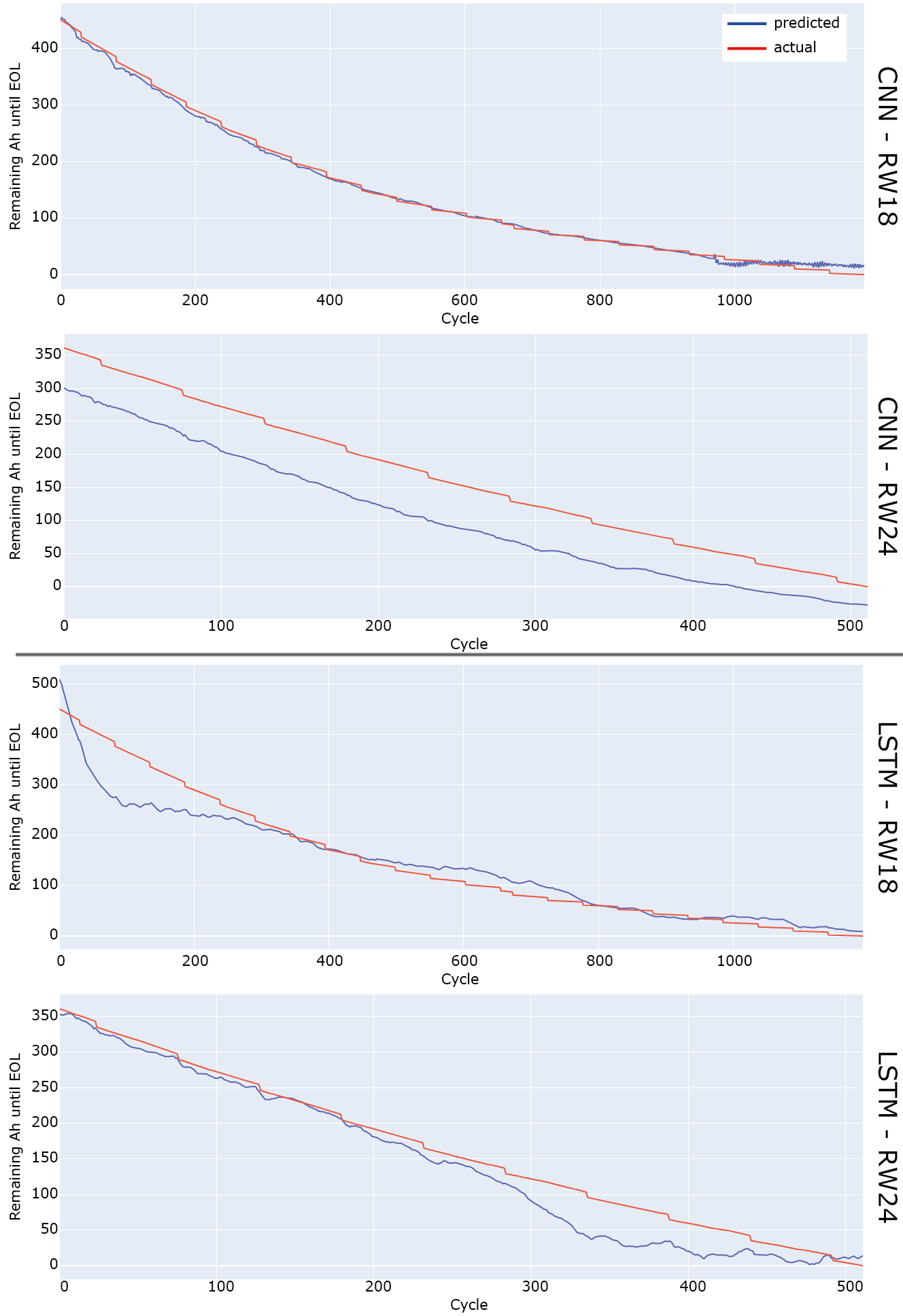}
 
    \caption{{A side-by-side example of the results of the CNN and the LSTM in ah-RUL prediction on the NASA Randomized dataset. The two examples show better stability and curve fitting of the CNN but a lower offset from the LSTM.}}
    \label{fig:nasa-results}
\end{figure}

The CNN-based and the LSTM-based ah-RUL predictors achieved comparable results. The CNN obtained an RMSE of 0.0799, while the LSTM attained a 0.074 RMSE. Figure \ref{fig:nasa-results} compares the testing results of the CNN and the LSTM, taking as an example two batteries from different groups: a battery from the skewed-high RW at room temperature and a battery from the skewed-low RW 
at a temperature of 40 degrees C. 
While the LSTM achieved slightly better results, the CNN provided more stable and consistent results over time. This is evident by the plots showing that the CNN prediction curves are always well-fitted, even though they could have an offset compared to the real curve. The LSTM, instead, does not suffer from the offset but has much more 
irregular 
curves. The plots for all the batteries are published in the project repository.
It is worth remarking that the remaining ampere-hour is predicted instead of the number of cycles. Considering that, and considering the very diverse conditions applied to the batteries, the results of both networks are excellent, although with different strengths, i.e., well-fitted curves with offsets versus no offset but with 
irregular 
curves.

\vspace{-8pt}

\subsection{UNIBO Powertools}

\begin{figure}[H]
    \begin{adjustwidth}{-\extralength}{0cm}
        \centering
    \includegraphics[width=0.73\linewidth]{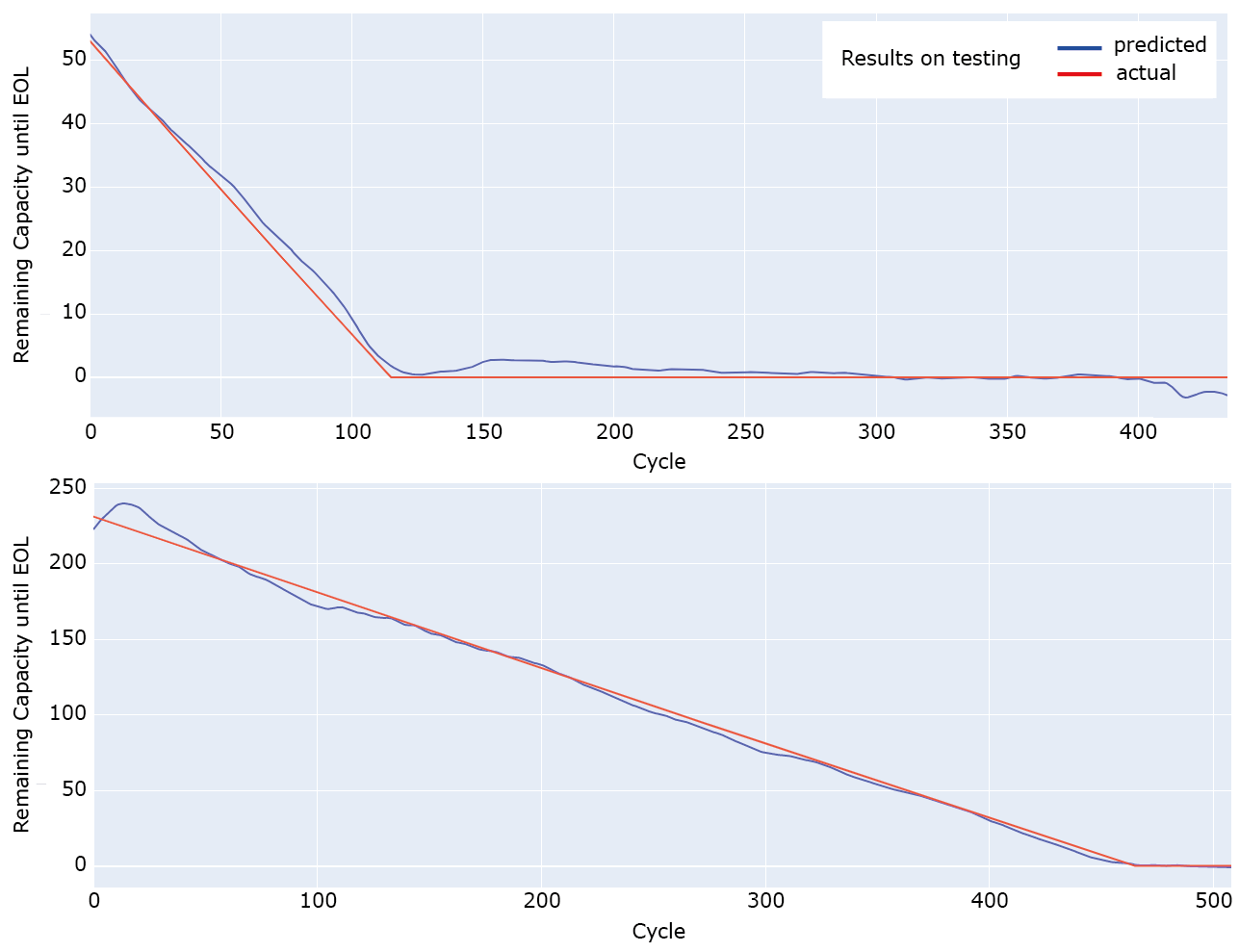}
    \end{adjustwidth}
    \caption{An example of test results of the LSTM ah-RUL prediction on the UNIBO Powertools~dataset.}
    \label{fig:unibo-results}
\end{figure}

In the UNIBO experiment, the batteries in the training and testing sets also came from all the group types to guarantee the fairness of the results. One battery from each of the 7 groups was selected for the testing set. All the other batteries were put in the training set. Batteries 047 and 049 were excluded as, at the time of the dataset construction, they were not cycled yet until end-of-life. Battery 019 was not used as some of its data are corrupted. This resulted in 7 batteries for the testing set and 20 for the training set.
The LSTM achieved an average RMSE of 0.021. An example of the prediction on the testing set is shown \mbox{in Figure \ref{fig:unibo-results}}. The results demonstrate the ability of the LSTM to learn the degradation trends of the batteries cycled under different conditions.

\section{Conclusions and Future Works}
\label{sec:conclusions}

In this paper, we proposed and compared two models using the NASA Randomized Battery Usage dataset for the RUL estimation of Li-ion batteries based on an autoencoder plus CNN and an autoencoder plus LSTM. In addition, an LSTM was proposed to estimate the RUL in the UNIBO Powertools dataset.

Aiming to push forward the applicability to real cases of the current deep-learning-based methods for RUL estimation, we proposed a novel definition of RUL based on ampere-hours, which is more useful for real scenarios. We also employed two datasets with a wide range of cycling conditions to ensure the generalization of the methods.
Compared to the datasets used in the literature so far, which employ a limited amount of batteries typically discharged under constant current (a condition that is not realistic in EVs), the NASA Randomized dataset provides an entirely different difficulty in predicting the RUL. The UNIBO Powertools dataset also provides a fresh perspective on data diversity as it contains batteries from various manufacturers and with different specifications.

To the best of the authors' knowledge, this is the first successful application of deep-learning-based methods for RUL estimation on such a vast number of Li-ion batteries cycled under different conditions, 
and the first study to use the UNIBO dataset. 
This is also the first application of those methods on an RUL that is not based on the simplified concept of cycles.
The results show that the 
particular 
 autoencoder employed can extract the dominant features of the cycle curves, and that both the CNN and the LSTM proposed can predict the RUL based on those features.

Several directions are open for future investigation. While the results obtained on such complex data are encouraging, there is still room for improvement. It would be interesting to design a network that provides the advantages of both the CNN (well-fitted curves) and the LSTM (no offset). In addition, transformers \cite{vaswani2017attention}, which have achieved impressive results in recent times, should be 
studied. This kind of NN can learn temporal dependencies that are even longer than the ones learnable by an LSTM. Therefore, it is quite promising for the objective. To the best of the authors' knowledge, this kind of NN has not been tested yet for RUL estimation. The transformer will substitute the current LSTM, in the hope of achieving even better performances.

The improvement of battery RUL estimation can support the development of battery recycling. Still, policies will be required to fully develop it, either with economic incentives \cite{recycling-tang} or with the involvement of governments \cite{recycling-hao}. Further work in this direction is \mbox{also advised.
}

\vspace{6pt}

\authorcontributions{Conceptualization, M.B. and C.F.; methodology, M.B.; software, M.B.; validation, M.B.; formal analysis, M.B.; investigation, M.B.; resources, G.P.; data curation, M.B.; writing---original draft preparation, M.B.; writing---review and editing, M.B.; visualization, M.B.; supervision, G.P. and C.R.; project administration, G.P.; funding acquisition, G.P. All authors have read and agreed to the published version of the manuscript.}

\funding{This research was funded by the LiBER project under the POR FESR program by Emilia Romagna Region, years 2019--2022.}

\dataavailability{Data are available from the sources cited in the article. The UNIBO Powertools dataset is available at \href{https://doi.org/10.17632/n6xg5fzsbv.1}{https://doi.org/10.17632/n6xg5fzsbv.1} and the Nasa Randomized Battery Usage dataset is available at \href{https://www.nasa.gov/content/prognostics-center-of-excellence-data-set-repository}{https://www.nasa.gov/content/prognostics-center-of-excellence-data-set-repository} {(accessed on 15 March 2023).} 
} 


\conflictsofinterest{The authors declare no conflict of interest.}

\begin{adjustwidth}{-\extralength}{0cm}

\reftitle{References}



\PublishersNote{}
\end{adjustwidth}

\begin{thebibliography}{999}

\bibitem[Hofmann \em{et~al.}(2016)Hofmann, Guan, Chalvatzis, and
Huo]{HOFMANN2016995}
Hofmann, J.; Guan, D.; Chalvatzis, K.; Huo, H.
\newblock Assessment of electrical vehicles as a successful driver for reducing
CO\textsubscript{2} emissions in China.
\newblock {\em Appl. Energy} {\bf 2016}, {\em 184},~995--1003.
\newblock
[\href{http://doi.org/10.1016/j.apenergy.2016.06.042}{CrossRef}]

\bibitem[Zou \em{et~al.}(2016)Zou, Wei, Sun, Hu, and Shiao]{ZOU201625}
Zou, Y.; Wei, S.; Sun, F.; Hu, X.; Shiao, Y.
\newblock Large-scale deployment of electric taxis in Beijing: A real-world
analysis.
\newblock {\em Energy} {\bf 2016}, {\em 100},~25--39. [\href{http://dx.doi.org/10.1016/j.energy.2016.01.062}{CrossRef}]

\bibitem[{US Environmental Protection Agency (EPA)}(2014)]{epa2014}
{US Environmental Protection Agency (EPA)}.
\newblock {Global Greenhouse Gas Emissions Data}.  2014.
\newblock {Available online:} 
\url{https://www.epa.gov/ghgemissions/global-greenhouse-gas-emissions-data} (accessed on 3 May 2021).

\bibitem[{European Commission}(2012)]{eu2012}
{European Commission}.
\newblock {Air Pollution from the Main Sources---Air Emissions from Road
Vehicles}.  2012.
\newblock {Available online:} 
\url{https://ec.europa.eu/environment/air/sources/road.htm} (accessed on 3 May 2021).

\bibitem[Kleeman \em{et~al.}(2000)Kleeman, Schauer, and
Cass]{es981276y}
Kleeman, M.J.; Schauer, J.J.; Cass, G.R.
\newblock Size and Composition Distribution of Fine Particulate Matter Emitted
from Motor Vehicles.
\newblock {\em Environ. Sci. Technol.} {\bf 2000}, {\em
34},~1132--1142. [\href{http://dx.doi.org/10.1021/es981276y}{CrossRef}]

\bibitem[Kheirbek \em{et~al.}(2016)Kheirbek, Haney, Douglas, Ito, and
Matte]{kheirbek2016contribution}
Kheirbek, I.; Haney, J.; Douglas, S.; Ito, K.; Matte, T.
\newblock The contribution of motor vehicle emissions to ambient fine
particulate matter public health impacts in New York City: A health burden
assessment.
\newblock {\em Environ. Health} {\bf 2016}, {\em 15},~89. [\href{http://dx.doi.org/10.1186/s12940-016-0172-6}{CrossRef}]

\bibitem[Koolen and Rothenberg(2019)]{koolen2019air}
Koolen, C.D.; Rothenberg, G.
\newblock Air pollution in Europe.
\newblock {\em ChemSusChem} {\bf 2019}, {\em 12},~164--172. [\href{http://dx.doi.org/10.1002/cssc.201802292}{CrossRef}]

\bibitem[Anderson \em{et~al.}(2005)Anderson, Atkinson, Peacock, Sweeting, and
Marston]{anderson2005ambient}
Anderson, H.R.; Atkinson, R.W.; Peacock, J.L.; Sweeting, M.J.; Marston, L.
\newblock Ambient particulate matter and health effects: Publication bias in
studies of short-term associations.
\newblock {\em Epidemiology} {\bf 2005}, {\em 16}, 155--163. [\href{http://dx.doi.org/10.1097/01.ede.0000152528.22746.0f}{CrossRef}] [\href{http://www.ncbi.nlm.nih.gov/pubmed/15703529}{PubMed}]

\bibitem[Brunekreef and Forsberg(2005)]{brunekreef2005epidemiological}
Brunekreef, B.; Forsberg, B.
\newblock Epidemiological evidence of effects of coarse airborne particles on
health.
\newblock {\em Eur. Respir. J.} {\bf 2005}, {\em 26},~309--318. [\href{http://dx.doi.org/10.1183/09031936.05.00001805}{CrossRef}]

\bibitem[Opitz \em{et~al.}(2017)Opitz, Badami, Shen, Vignarooban, and
Kannan]{OPITZ2017685}
Opitz, A.; Badami, P.; Shen, L.; Vignarooban, K.; Kannan, A.
\newblock Can Li-Ion batteries be the panacea for automotive applications?
\newblock {\em Renew. Sustain. Energy Rev.} {\bf 2017}, {\em
68},~685--692. [\href{http://dx.doi.org/10.1016/j.rser.2016.10.019}{CrossRef}]

\bibitem[Manzetti and Mariasiu(2015)]{MANZETTI20151004}
Manzetti, S.; Mariasiu, F.
\newblock Electric vehicle battery technologies: From present state to future
systems.
\newblock {\em Renew. Sustain. Energy Rev.} {\bf 2015}, {\em
51},~1004--1012. [\href{http://dx.doi.org/10.1016/j.rser.2015.07.010}{CrossRef}]

\bibitem[Hannan \em{et~al.}(2017)Hannan, Lipu, Hussain, and
Mohamed]{HANNAN2017834}
Hannan, M.; Lipu, M.; Hussain, A.; Mohamed, A.
\newblock A review of lithium-ion battery state of charge estimation and
management system in electric vehicle applications: Challenges and
recommendations.
\newblock {\em Renew. Sustain. Energy Rev.} {\bf 2017}, {\em
78},~834--854. [\href{http://dx.doi.org/10.1016/j.rser.2017.05.001}{CrossRef}]

\bibitem[Zhang \em{et~al.}(2019)Zhang, Zhai, Guo, Wang, Peng, and
Zhang]{ZHANG2019100951}
Zhang, S.; Zhai, B.; Guo, X.; Wang, K.; Peng, N.; Zhang, X.
\newblock Synchronous estimation of state of health and remaining useful
lifetime for lithium-ion battery using the incremental capacity and
artificial neural networks.
\newblock {\em J. Energy Storage} {\bf 2019}, {\em 26},~100951. [\href{http://dx.doi.org/10.1016/j.est.2019.100951}{CrossRef}]

\bibitem[LIU \em{et~al.}(2014)LIU, ZHOU, and PENG]{liu2014data}
Liu, D.; Zhou, J.; Peng, Y.
\newblock Data-driven prognostics and remaining useful life estimation for
lithium-ion battery: A review.
\newblock {\em Instrumentation} {\bf 2014}, {\em {1}, 59--70. 
}


\bibitem[Hu \em{et~al.}(2017)Hu, Cao, and Egardt]{hu2017condition}
Hu, X.; Cao, D.; Egardt, B.
\newblock Condition monitoring in advanced battery management systems: Moving
horizon estimation using a reduced electrochemical model.
\newblock {\em IEEE/ASME Trans. Mechatronics} {\bf 2017}, {\em
23},~167--178. [\href{http://dx.doi.org/10.1109/TMECH.2017.2675920}{CrossRef}]

\bibitem[Williard \em{et~al.}(2013)Williard, He, Hendricks, and
Pecht]{en6094682}
Williard, N.; He, W.; Hendricks, C.; Pecht, M.
\newblock Lessons Learned from the 787 Dreamliner Issue on Lithium-Ion Battery
Reliability.
\newblock {\em Energies} {\bf 2013}, {\em 6},~4682--4695. [\href{http://dx.doi.org/10.3390/en6094682}{CrossRef}]

\bibitem[NASA(2007)]{nasa2007}
NASA.
\newblock \textit{Mars Global Surveyor (MGS) Spacecraft Loss of Contact};{ NASA: Washington, DC, USA,} 
2007.

\bibitem[Valdes-Dapena and Peter(2021)]{ValdesDapena2021May}
Isidore, C.; Valdes-Dapena, P.
\newblock {Hyundai's Recall of 82,000 Electric Cars is One of the Most
Expensive in History}.  2021.
\newblock {Available online:} 
\url{https://edition.cnn.com/2021/02/25/tech/hyundai-ev-recall/index.html} (accessed on 3 May 2021).

\bibitem[Hawkins(2020)]{Hawkins2020Nov}
Hawkins, A.J.
\newblock {GM Recalls 68,000 Electric Chevy Bolts over Battery Fire Concerns}.
\newblock {\em Verge} {\bf {2020}}. 
\newblock {Available online:}
\url{https://www.theverge.com/2020/11/13/21564217/gm-chevy-bolt-recall-battery-fire-lg-chem} {(accessed on 15 March 2023).}

\bibitem[Bilgin \em{et~al.}(2015)Bilgin, Magne, Malysz, Yang, Pantelic,
Preindl, Korobkine, Jiang, Lawford, and Emadi]{7112507}
Bilgin, B.; Magne, P.; Malysz, P.; Yang, Y.; Pantelic, V.; Preindl, M.;
Korobkine, A.; Jiang, W.; Lawford, M.; Emadi, A.
\newblock Making the Case for Electrified Transportation.
\newblock {\em IEEE Trans. Transp. Electrif.} {\bf 2015},
{\em 1},~4--17. [\href{http://dx.doi.org/10.1109/TTE.2015.2437338}{CrossRef}]

\bibitem[{Mahmoudzadeh Andwari} \em{et~al.}(2017){Mahmoudzadeh Andwari},
Pesiridis, Rajoo, Martinez-Botas, and Esfahanian]{MAHMOUDZADEHANDWARI2017414}
{Mahmoudzadeh Andwari}, A.; Pesiridis, A.; Rajoo, S.; Martinez-Botas, R.;
Esfahanian, V.
\newblock A review of Battery Electric Vehicle technology and readiness levels.
\newblock {\em Renew. Sustain. Energy Rev.} {\bf 2017}, {\em
78},~414--430. [\href{http://dx.doi.org/10.1016/j.rser.2017.03.138}{CrossRef}]

\bibitem[Wu \em{et~al.}(2020)Wu, Xue, Shen, Lei, Chen, and Liu]{8986572}
Wu, Y.; Xue, Q.; Shen, J.; Lei, Z.; Chen, Z.; Liu, Y.
\newblock State of Health Estimation for Lithium-Ion Batteries Based on Healthy
Features and Long Short-Term Memory.
\newblock {\em IEEE Access} {\bf 2020}, {\em 8},~28533--28547. [\href{http://dx.doi.org/10.1109/ACCESS.2020.2972344}{CrossRef}]

\bibitem[Johnson(2014)]{JOHNSON2014582}
Johnson, N.
\newblock 19---Battery technology for CO\textsubscript{2} reduction. In {\em Alternative Fuels
and Advanced Vehicle Technologies for Improved Environmental Performance};
Folkson, R., Ed.; Woodhead Publishing: Sawston, UK, 2014; pp. 582--631. [\href{http://dx.doi.org/10.1533/9780857097422.3.582}{CrossRef}]

\bibitem[Barré \em{et~al.}(2013)Barré, Deguilhem, Grolleau, Gérard, Suard,
and Riu]{BARRE2013680}
Barré, A.; Deguilhem, B.; Grolleau, S.; Gérard, M.; Suard, F.; Riu, D.
\newblock A review on lithium-ion battery ageing mechanisms and estimations for
automotive applications.
\newblock {\em J. Power Sources} {\bf 2013}, {\em 241},~680--689.
\newblock
[\href{http://dx.doi.org/10.1016/j.jpowsour.2013.05.040}{CrossRef}]

\bibitem[Xiong \em{et~al.}(2018)Xiong, Li, and Tian]{XIONG201818}
Xiong, R.; Li, L.; Tian, J.
\newblock Towards a smarter battery management system: A critical review on
battery state of health monitoring methods.
\newblock {\em J. Power Sources} {\bf 2018}, {\em 405},~18--29.
\newblock
[\href{http://dx.doi.org/10.1016/j.jpowsour.2018.10.019}{CrossRef}]

\bibitem[Lu \em{et~al.}(2013)Lu, Han, Li, Hua, and Ouyang]{LU2013272}
Lu, L.; Han, X.; Li, J.; Hua, J.; Ouyang, M.
\newblock A review on the key issues for lithium-ion battery management in
electric vehicles.
\newblock {\em J. Power Sources} {\bf 2013}, {\em 226},~272--288.
\newblock
[\href{http://dx.doi.org/10.1016/j.jpowsour.2012.10.060}{CrossRef}]

\bibitem[Fan \em{et~al.}(2020)Fan, Xiao, Li, Yang, and Tang]{FAN2020101741}
Fan, Y.; Xiao, F.; Li, C.; Yang, G.; Tang, X.
\newblock A novel deep learning framework for state of health estimation of
lithium-ion battery.
\newblock {\em J. Energy Storage} {\bf 2020}, {\em 32},~101741. [\href{http://dx.doi.org/10.1016/j.est.2020.101741}{CrossRef}]

\bibitem[Vidal \em{et~al.}(2020)Vidal, Malysz, Kollmeyer, and Emadi]{9036949}
Vidal, C.; Malysz, P.; Kollmeyer, P.; Emadi, A.
\newblock Machine Learning Applied to Electrified Vehicle Battery State of
Charge and State of Health Estimation: State-of-the-Art.
\newblock {\em IEEE Access} {\bf 2020}, {\em 8},~52796--52814. [\href{http://dx.doi.org/10.1109/ACCESS.2020.2980961}{CrossRef}]





\bibitem{IEC}
\textit{IEC-62660-2};
\newblock Secondary Lithium-Ion Cells for the Propulsion of Electric Road Vehicles---Part 2: Reliability and Abuse Testing 2018.
\newblock International Electrotechnical Commission: Geneva, Switzerland, 2018.

\bibitem{ISO}
\textit{ISO 12405-3};
\newblock ISO Electrically Propelled Road Vehicles---Test Specification for Lithium-Ion Traction Battery Packs and \mbox{Systems---Part} 3: Safety Performance Requirements. ISO: Geneva, Switzerland, 2018.

\bibitem{IEEE-standard}
\textit{IEEE Std 450-2020 (Revision IEEE Std 450-2010)};
\newblock IEEE Recommended Practice for Maintenance, Testing, and Replacement of
Vented Lead-Acid Batteries for Stationary Applications. IEEE: New York, NY, USA, 2021; pp. 1--71.






\bibitem[Duong(2000)]{DUONG2000244}
Duong, T.Q.
\newblock USABC and PNGV test procedures.
\newblock {\em J. Power Sources} {\bf 2000}, {\em 89},~244--248.
\newblock
[\href{http://dx.doi.org/10.1016/S0378-7753(00)00439-0}{CrossRef}]

\bibitem[Xing \em{et~al.}(2013)Xing, Ma, Tsui, and Pecht]{calce}
Xing, Y.; Ma, E.W.; Tsui, K.L.; Pecht, M.
\newblock An ensemble model for predicting the remaining useful performance of
lithium-ion batteries.
\newblock {\em Microelectron. Reliab.} {\bf 2013}, {\em 53},~811--820.
\newblock
[\href{http://dx.doi.org/10.1016/j.microrel.2012.12.003}{CrossRef}]



\bibitem[Saha and Goebel(2007) Saha, and Goebel]{nasadata}
Saha, B.; Goebel, K.
\newblock Battery Data Set.
\newblock {NASA Ames Prognostics Data Repository}. { 2007}.
\newblock {Available online:}
\url{https://www.nasa.gov/content/prognostics-center-of-excellence-data-set-repository} {(accessed on 15 March 2023).}


\bibitem[Chen \em{et~al.}(2020)Chen, He, Li, Chen, and
Zhang]{chen2020remaining}
Chen, Y.; He, Y.; Li, Z.; Chen, L.; Zhang, C.
\newblock Remaining Useful Life Prediction and State of Health Diagnosis of
Lithium-Ion Battery Based on Second-Order Central Difference Particle Filter.
\newblock {\em IEEE Access} {\bf 2020}, {\em 8},~37305--37313. [\href{http://dx.doi.org/10.1109/ACCESS.2020.2974401}{CrossRef}]

\bibitem[Li \em{et~al.}(2017)Li, Shu, Shen, Xiao, Yan, and Chen]{en10050691}
Li, X.; Shu, X.; Shen, J.; Xiao, R.; Yan, W.; Chen, Z.
\newblock An On-Board Remaining Useful Life Estimation Algorithm for
Lithium-Ion Batteries of Electric Vehicles.
\newblock {\em Energies} {\bf 2017}, {\em 10}, 691. [\href{http://dx.doi.org/10.3390/en10050691}{CrossRef}]

\bibitem[Ng \em{et~al.}(2014)Ng, Xing, and Tsui]{NG2014114}
Ng, S.S.; Xing, Y.; Tsui, K.L.
\newblock A naive Bayes model for robust remaining useful life prediction of
lithium-ion battery.
\newblock {\em Appl. Energy} {\bf 2014}, {\em 118},~114--123.
\newblock
[\href{http://dx.doi.org/10.1016/j.apenergy.2013.12.020}{CrossRef}]

\bibitem[Hu \em{et~al.}(2019)Hu, Feng, Liu, Zhang, Xie, and Liu]{HU2019109334}
Hu, X.; Feng, F.; Liu, K.; Zhang, L.; Xie, J.; Liu, B.
\newblock State estimation for advanced battery management: Key challenges and
future trends.
\newblock {\em Renew. Sustain. Energy Rev.} {\bf 2019}, {\em
114},~109334. [\href{http://dx.doi.org/10.1016/j.rser.2019.109334}{CrossRef}]

\bibitem[Song \em{et~al.}(2020)Song, Zhang, Liang, Han, and
Zhang]{SONG2020101836}
Song, L.; Zhang, K.; Liang, T.; Han, X.; Zhang, Y.
\newblock Intelligent state of health estimation for lithium-ion battery pack
based on big data analysis.
\newblock {\em J. Energy Storage} {\bf 2020}, {\em 32},~101836. [\href{http://dx.doi.org/10.1016/j.est.2020.101836}{CrossRef}]

\bibitem[Venugopal and T.(2019)]{en12224338}
Venugopal, P.; Vigneswaran, T.
\newblock State-of-Health Estimation of Li-ion Batteries in Electric Vehicle
Using IndRNN under Variable Load Condition.
\newblock {\em Energies} {\bf 2019}, {\em 12}, 4338. [\href{http://dx.doi.org/10.3390/en12224338}{CrossRef}]

\bibitem[Ahmadi \em{et~al.}(2014)Ahmadi, Yip, Fowler, Young, and
Fraser]{AHMADI201464}
Ahmadi, L.; Yip, A.; Fowler, M.; Young, S.B.; Fraser, R.A.
\newblock Environmental feasibility of re-use of electric vehicle batteries.
\newblock {\em Sustain. Energy Technol. Assess.} {\bf 2014},
{\em 6},~64--74. [\href{http://dx.doi.org/10.1016/j.seta.2014.01.006}{CrossRef}]

\bibitem[Shen \em{et~al.}(2019)Shen, Sadoughi, Chen, Hong, and
Hu]{SHEN2019100817}
Shen, S.; Sadoughi, M.; Chen, X.; Hong, M.; Hu, C.
\newblock A deep learning method for online capacity estimation of lithium-ion
batteries.
\newblock {\em J. Energy Storage} {\bf 2019}, {\em 25},~100817. [\href{http://dx.doi.org/10.1016/j.est.2019.100817}{CrossRef}]

\bibitem[Harper \em{et~al.}(2019)Harper, Sommerville, Kendrick, Driscoll,
Slater, Stolkin, Walton, Christensen, Heidrich, Lambert,
et~al.]{harper2019recycling}
Harper, G.; Sommerville, R.; Kendrick, E.; Driscoll, L.; Slater, P.; Stolkin,
R.; Walton, A.; Christensen, P.; Heidrich, O.; \linebreak Lambert, S.;  et~al.
\newblock Recycling lithium-ion batteries from electric vehicles.
\newblock {\em Nature} {\bf 2019}, {\em 575},~75--86. [\href{http://dx.doi.org/10.1038/s41586-019-1682-5}{CrossRef}]




\bibitem[Pražanová \em{et~al.}(2022)Pražanová, Knap, Vaclav, and Stroe]{prazanova}
Pražanová, A.; Knap, V.; Stroe, D.-I.
\newblock Literature Review, Recycling of Lithium-Ion Batteries from Electric Vehicles, Part II: Environmental and Economic Perspective.
\newblock {\em Energies} {\bf 2022}, {\em 15}, 7356. [\href{http://dx.doi.org/10.3390/en15197356}{CrossRef}]




\bibitem[Lipu \em{et~al.}(2018)Lipu, Hannan, Hussain, Hoque, Ker, Saad, and
Ayob]{LIPU2018115}
Lipu, M.H.; Hannan, M.; Hussain, A.; Hoque, M.; Ker, P.J.; Saad, M.; Ayob, A.
\newblock A review of state of health and remaining useful life estimation
methods for lithium-ion battery in electric vehicles: Challenges and
recommendations.
\newblock {\em J. Clean. Prod.} {\bf 2018}, {\em 205},~115--133.
\newblock
[\href{http://dx.doi.org/10.1016/j.jclepro.2018.09.065}{CrossRef}]

\bibitem[Dai \em{et~al.}(2019)Dai, Zhao, Lin, Wu, and Zheng]{8536873}
Dai, H.; Zhao, G.; Lin, M.; Wu, J.; Zheng, G.
\newblock A Novel Estimation Method for the State of Health of Lithium-Ion
Battery Using Prior Knowledge-Based Neural Network and Markov Chain.
\newblock {\em IEEE Trans. Ind. Electron.} {\bf 2019}, {\em
66},~7706--7716.   [\href{http://dx.doi.org/10.1109/TIE.2018.2880703}{CrossRef}]

\bibitem[Krizhevsky \em{et~al.}(2012)Krizhevsky, Sutskever, and
Hinton]{NIPS2012_4824}
Krizhevsky, A.; Sutskever, I.; Hinton, G.E.
\newblock ImageNet Classification with Deep Convolutional Neural Networks. In
{\em Advances in Neural Information Processing Systems 25}; Pereira, F.,
Burges, C.J.C., Bottou, L., Weinberger, K.Q., Eds.; Curran Associates, Inc.: Red Hook, NY, USA, 2012; pp. 1097--1105.

\bibitem[Hametner \em{et~al.}(2016)Hametner, Jakubek, and Prochazka]{7811778}
Hametner, C.; Jakubek, S.; Prochazka, W.
\newblock Data-driven design of a cascaded observer for battery state of health
estimation.
\newblock In Proceedings of the 2016 IEEE International Conference on
Sustainable Energy Technologies (ICSET), Hanoi, Vietnam, 14--16 November 2016; pp. 180--185. [\href{http://dx.doi.org/10.1109/ICSET.2016.7811778}{CrossRef}]

\bibitem[Hochreiter and Schmidhuber(1997)]{hochreiter1997long}
Hochreiter, S.; Schmidhuber, J.
\newblock Long short-term memory.
\newblock {\em Neural Comput.} {\bf 1997}, {\em 9},~1735--1780. [\href{http://dx.doi.org/10.1162/neco.1997.9.8.1735}{CrossRef}]

\bibitem[{dos Reis} \em{et~al.}(2021){dos Reis}, Strange, Yadav, and
Li]{DOSREIS2021100081}
{Dos Reis}, G.; Strange, C.; Yadav, M.; Li, S.
\newblock Lithium-ion battery data and where to find it.
\newblock {\em Energy AI} {\bf 2021}, \textit{5}, 100081.\linebreak
\newblock. [\href{http://dx.doi.org/10.1016/j.egyai.2021.100081}{CrossRef}]

\bibitem[Bole \em{et~al.}(2014)Bole, Kulkarni, and Daigle]{bole2014randomized}
Bole, B.; Kulkarni, C.; Daigle, M.
\newblock Randomized Battery Usage Data Set.
\newblock { NASA Ames Prognostics Data Repository}. {2014}.
\newblock{Available online:}
\url{https://www.nasa.gov/content/prognostics-center-of-excellence-data-set-repository} {(accessed on 15 March 2023).}


\bibitem[Ungurean \em{et~al.}(2020)Ungurean, Micea, and
Cârstoiu]{onlineestimationrnn2020}
Ungurean, L.; Micea, M.V.; Cârstoiu, G.
\newblock Online state of health prediction method for lithium-ion batteries,
based on gated recurrent unit neural networks.
\newblock {\em Int. J. Energy Res.} {\bf 2020}, {\em
44},~6767--6777. [\href{http://dx.doi.org/10.1002/er.5413}{CrossRef}]

\bibitem[Zhou \em{et~al.}(2020)Zhou, Li, Zhu, Zhang, and Hou]{9037249}
Zhou, D.; Li, Z.; Zhu, J.; Zhang, H.; Hou, L.
\newblock State of Health Monitoring and Remaining Useful Life Prediction of\linebreak
Lithium-Ion Batteries Based on Temporal Convolutional Network.
\newblock {\em IEEE Access} {\bf 2020}, {\em 8},~53307--53320. [\href{http://dx.doi.org/10.1109/ACCESS.2020.2981261}{CrossRef}]

\bibitem[Zhang \em{et~al.}(2017)Zhang, Xiong, He, and Liu]{8079316}
Zhang, Y.; Xiong, R.; He, H.; Liu, Z.
\newblock A LSTM-RNN method for the lithuim-ion battery remaining useful life
prediction.
\newblock \linebreak In Proceedings of the 2017 Prognostics and System Health Management
Conference (PHM-Harbin), Harbin, China, \linebreak 9--12 July 2017; pp. 1--4. [\href{http://dx.doi.org/10.1109/PHM.2017.8079316}{CrossRef}]

\bibitem[Zhang \em{et~al.}(2020)Zhang, Li, and Li]{ZHANG2020108052}
Zhang, W.; Li, X.; Li, X.
\newblock Deep learning-based prognostic approach for lithium-ion batteries
with adaptive time-series prediction and on-line validation.
\newblock {\em Measurement} {\bf 2020}, {\em 164},~108052.
\newblock
[\href{http://dx.doi.org/10.1016/j.measurement.2020.108052}{CrossRef}]

\bibitem[Li \em{et~al.}(2020)Li, Zhang, Xiong, Ding, Hou, Luo, Rong, and
Li]{LI2020228069}
Li, P.; Zhang, Z.; Xiong, Q.; Ding, B.; Hou, J.; Luo, D.; Rong, Y.; Li, S.
\newblock State-of-health estimation and remaining useful life prediction for
the lithium-ion battery based on a variant long short term memory neural
network.
\newblock {\em J. Power Sources} {\bf 2020}, {\em 459},~228069.
\newblock
[\href{http://dx.doi.org/10.1016/j.jpowsour.2020.228069}{CrossRef}]

\bibitem[Ren \em{et~al.}(2018)Ren, Zhao, Hong, Zhao, Wang, and Zhang]{8418374}
Ren, L.; Zhao, L.; Hong, S.; Zhao, S.; Wang, H.; Zhang, L.
\newblock Remaining Useful Life Prediction for Lithium-Ion Battery: A Deep
Learning Approach.
\newblock {\em IEEE Access} {\bf 2018}, {\em 6},~50587--50598. [\href{http://dx.doi.org/10.1109/ACCESS.2018.2858856}{CrossRef}]

\bibitem[Ren \em{et~al.}(2021)Ren, Dong, Wang, Meng, Zhao, and Deen]{9137406}
Ren, L.; Dong, J.; Wang, X.; Meng, Z.; Zhao, L.; Deen, M.J.
\newblock A Data-Driven Auto-CNN-LSTM Prediction Model for Lithium-Ion Battery
Remaining Useful Life.
\newblock {\em IEEE Trans. Ind. Inform.} {\bf 2021}, {\em
17},~3478--3487. [\href{http://dx.doi.org/10.1109/TII.2020.3008223}{CrossRef}]

\bibitem[Wong \em{et~al.}(2021)Wong, Bosello, Tse, Falcomer, Rossi, and
Pau]{10.1145/3462203.3475878}
Wong, K.L.; Bosello, M.; Tse, R.; Falcomer, C.; Rossi, C.; Pau, G.
\newblock Li-Ion Batteries State-of-Charge Estimation Using Deep LSTM at
Various Battery Specifications and Discharge Cycles.
\newblock In Proceedings of the Conference on Information
Technology for Social Good, GoodIT '21, Rome, Italy, 9--11 September 2021; Association for Computing Machinery: New York,
NY, USA, 2021; \linebreak pp. 85–90. [\href{http://dx.doi.org/10.1145/3462203.3475878}{CrossRef}]

\bibitem[Maggipinto \em{et~al.}(2018)Maggipinto, Masiero, Beghi, and
Susto]{MAGGIPINTO2018126}
Maggipinto, M.; Masiero, C.; Beghi, A.; Susto, G.A.
\newblock A Convolutional Autoencoder Approach for Feature Extraction in
Virtual Metrology.
\newblock {\em {Procedia Manuf.}} {\bf 2018}, {\em 17},~126--133.

\bibitem[Vaswani \em{et~al.}(2017)Vaswani, Shazeer, Parmar, Uszkoreit, Jones,
Gomez, Kaiser, and Polosukhin]{vaswani2017attention}
Vaswani, A.; Shazeer, N.; Parmar, N.; Uszkoreit, J.; Jones, L.; Gomez, A.N.;
Kaiser, L.; Polosukhin, I.
\newblock Attention Is All You Need. \textit{arXiv } \textbf{2017}, arXiv:1706.03762.










\bibitem[Tang \em{et~al.}(2018)Tang, Zhang, Li, Wang, and
Li]{recycling-tang}
Tang, Y.; Zhang, Q.; Li, Y.; Wang, G.; Li, Y.
\newblock Recycling mechanisms and policy suggestions for spent electric vehicles' power battery---A case of Beijing.
\newblock {\em J. Clean. Prod.} {\bf 2018}, {\em 186}, 388--406. [\href{http://dx.doi.org/10.1016/j.jclepro.2018.03.043}{CrossRef}]


\bibitem[Hao \em{et~al.}(2022)Hao, Xu, Wei, Wu, and
Xu]{recycling-hao}
Hao, H.; Xu, W.; Wei, F.; Wu, C.; Xu, Z.
\newblock Reward-Penalty vs. Deposit-Refund: Government Incentive Mechanisms for EV Battery Recycling.
\newblock {\em Energies} {\bf 2022}, {\em 15}, 6885. [\href{http://dx.doi.org/10.3390/en15196885}{CrossRef}]










\end{thebibliography}
\end{document}